\documentclass[10pt,journal,compsoc]{IEEEtran}

\ifCLASSOPTIONcompsoc
  \usepackage[nocompress]{cite}
\else
  \usepackage{cite}
\fi
\ifCLASSINFOpdf
  \usepackage[pdftex]{graphicx}
  \graphicspath{{images/}{../jpeg/}{../pdf/}}
  \DeclareGraphicsExtensions{.pdf,.jpeg,.png}
\else
  \usepackage[dvips]{graphicx}
  \graphicspath{images/}{../eps/}}
  \DeclareGraphicsExtensions{.eps}
\fi
\usepackage{amsmath}
\begin{document}
%
\title{Image Super-Resolution Using \\ VDSR-ResNeXt and SRCGAN}
%
%
%
%

\author{Saifuddin~Hitawala,
        Yao~Li,
        Xian~Wang
        and~Dongyang~Yang \\ Computer Science and Statistics Department, University of Waterloo \\ \{saifuddin.hitawala, y547li, x249wang, d39yang\}@uwaterloo.ca}
\maketitle


%

\section{Introduction}
%
%
%
%
Image super-resolution (SR) is a computer vision task involving reconstruction of high-resolution (HR) images given a low-resolution (LR) image as input. It is highly challenging and has many practical applications, such as medical image processing\cite{Med_image_processing}, satellite imaging \cite{Satellite_Imaging}, facial image  enhancement \cite{facial_image, facial_image_enhancement} and compressed image improvement \cite{compressed_image_improvement}. 

Over the past decade, many SR techniques have been developed using deep learning. Among those, generative adversarial networks (GAN) \cite{GAN} and very deep convolutional networks (VDSR) \cite{vdsr_paper} have shown promising results in terms of HR image quality and computational speed. In this paper, we propose two approaches based on these two algorithms: VDSR-ResNeXt, which is a deep multi-branch convolutional network inspired by VDSR and ResNeXt; and SRCGAN, which is a conditional GAN that explicitly passes class labels as input to the GAN. The two methods were implemented on common SR benchmark datasets for both quantitative and qualitative assessment.

\section{Related Work}

There is a vast amount of literature on single image super-resolution using deep learning techniques. \textbf{SRCNN} \cite{srcnn_paper} is one of the pioneers that demonstrated the potential for convolutional networks to be applied to this research domain, achieving then state-of-art reconstruction quality while maintaining a lightweight structure. The model consists of 3 convolutional layers, each responsible for feature extraction, non-linear mapping to high-resolution patch representations and reconstruction, respectively. Although SRCNN learns an end-to-end mapping between LR and HR images, it is still a fairly shallow network by current standards, and superior performances have been reported with deeper models and other CNN architecture variants. Moreover, since SRCNN can be trained only on a single scale of magnification, it is constrained to work only on the specific scale on which it has been trained. This led us to consider other techniques. 

\vspace{0.5em}

\textbf{VDSR} \cite{vdsr_paper} is one of the fastest models while also capable of providing highly accurate image construction. It is inspired by the VGG-net \cite{vgg}, and is made efficient with residual-learning (i.e. given ground truth HR image \textbf{y} and LR input \textbf{x}, predict on \textbf{r} = \textbf{y} - \textbf{x}, which makes more input values close to zero and more features zero after activation) and gradient clipping (restrict gradients to within a certain range to speed up convergence). VDSR stacks 18 convolutional layers with ReLU activation, in addition to an input convolutional layer that operates on the input image and an output convolutional layer used for image reconstruction. The receptive field of the network increases proportionally to the depth of the network, allowing the network to use more and more neighbor pixels to predict image details more accurately.

\vspace{0.5em}

\textbf{ResNeXt} \cite{resnext_paper} inspired by ResNet \cite{resnet_paper} is a model for image classification that is highly modularized. It involves stacking a series of homogeneous, multi-branch residual blocks, somewhat similar to ensemble learning except that training takes place jointly, not independently. The branches each perform their set of convolutions and non-linear transformations, and are then aggregated via summation at the end of the block. ResNeXt was shown to outperform ResNet for the task of image classification while maintaining the same model complexity. Due to their superiority in their respective tasks, ResNeXt and VDSR form the basis of our first proposed model - VDSR-ResNeXt.

\vspace{0.5em}

\textbf{SRGAN}\cite{srgan} represents another promising direction for image super-resolution using GANs. SRGAN uses a deep residual network as the generator to generate HR images from downscaled input. The combination of adversarial loss and content loss as the loss function also allows the model to generate high-resolution images towards the natural image manifold. Our second model is inspired from SRGAN as well as Conditional GANs because we feel the generative nature of the architecture fits naturally with our super-resolution task and GANs have shown competitive results in computer vision related research problems.

\section{Proposed Methods}

\subsection{VDSR-ResNeXt}
Inspired by the architecture of ResNeXt, we applied the idea of the multi-branch design to VDSR and named our model VDSR-ResNeXt. Our network is made up of a series of blocks sharing the same hyper-parameters (filter size, number of filters, number of layers). As in the case with ResNeXt, there are two equivalent designs for the building blocks: (i) 32 branches of 3-layer convolutional layers that are aggregated via summation at the end of the block (Fig. 1, middle), and (ii) grouped convolution operation performed on the middle layer of the block, in which the input channels are split into 32 groups for separate convolutions, and then concatenated as the outputs of the layer (Fig. 1, right). In practice, we used the grouped convolutions approach available with PyTorch’s \cite{pytorch} torch.nn.Conv2d function for more concise code. 

\begin{figure}[h]
\centering
\includegraphics[scale=0.43]{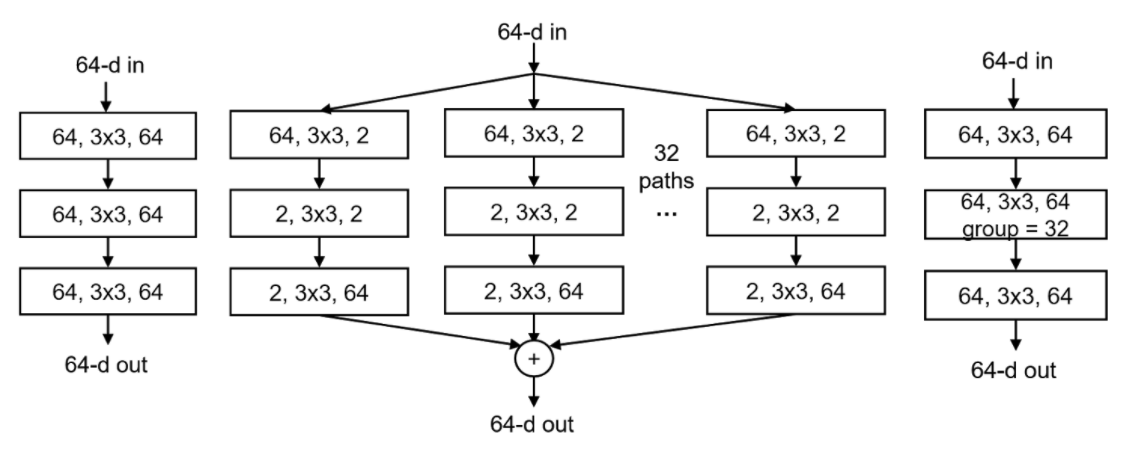}
\caption{(\textbf{left}) 3-layer snippet of the original VDSR architecture. (\textbf{middle}) Our proposed building block with multi-branch design. (\textbf{right}) Same as (middle), implemented as grouped convolutions.}
\end{figure}

In the original VDSR architecture, there are 18 $3\times3$ filter convolutional layers like the ones shown in Fig. 1 (left), which translates into 6 of our proposed 3-layer blocks if we keep the total number of layers the same. 

We used the mean squared error as the loss function:
\begin{equation}
\frac{1}{2}||\mathrm{y} - f(\mathrm{x})||^2
\end{equation} 
where $f(\mathrm{x})$ is the predicted image and $y$ is the target ground truth. The loss is averaged over the training set.

By adopting the split-transform-merge strategy, we were able to reduce the number of parameters and thus the model size, as shown in Table 1.

\begin{table}
\caption{Number of parameters with and without branches}
\centering
 \begin{tabular}{|c|c|c|} 
 \hline
 3 Conv. Layer & No. of Parameters & No. of Parameters \\ Configuration & Without Branches & With 32 Branches \\ [0.5ex] 
 \hline
 64, 3 x 3, 64  & \(\sim\)111K  & \(\sim\)75K \\ 
 64, 3 x 3, 64  & (= 64*3*3*64 & (= 32*(64*3*3*2 \\ 
 64, 3 x 3, 64  & + 64*3*3*64 & + 2*3*3*2 \\ 
  & + 64*3*3*64) & + 2*3*3*64)) \\ 
 \hline
 64, 3 x 3, 128  & \(\sim\)295K  & \(\sim\)152K \\ 
 128, 3 x 3, 128  & (= 64*3*3*128 & (= 32*(64*3*3*4 \\ 
 128, 3 x 3, 64  & + 128*3*3*128 & + 4*3*3*4 \\ 
  & + 128*3*3*64) & + 4*3*3*64)) \\ 
 \hline
 64, 3 x 3, 256  & \(\sim\)885K  & \(\sim\)313K \\ 
 256, 3 x 3, 256  & (= 64*3*3*256 & (= 32*(64*3*3*8 \\ 
 256, 3 x 3, 64  & + 256*3*3*256 & + 8*3*3*8 \\ 
  & + 256*3*3*64) & + 8*3*3*64)) \\  [1ex] 
 \hline
\end{tabular}
\end{table}
\setlength{\textfloatsep}{1pt plus 0.5pt minus 0.5pt}

We experimented with different configurations for our VDSR-ResNeXt architecture - increasing the number of channels within the block and varying the number of blocks. We experimented with the following configurations: VDSR-ResNeXt-18-64, VDSR-ResNeXt-18-128, VDSR-ResNeXt-18-256 and VDSR-ResNeXt-9-256, where the first number denotes the number of layers (excluding first and last layers), and the second number denotes the number of in and out channels for the layer on which the grouped convolution is applied. Due to page limits, we highlight in this report the results of VDSR-ResNeXt-18-128 and VDSR-ResNeXt-18-256, the two best performing configurations.

\subsection{SRCGAN}
\subsubsection{Generative Adversarial Nets (GAN)}

GANs \cite{GAN} have been considerably successful as a framework for generative models in recent years. The GAN architecture consists of two models: a generative model $G$ that captures the data distribution, and a discriminative model $D$ that tries to distinguish whether a sample is a generated output or the real data. Here, the generator and the discriminator are trained as part of a double feedback loop, so that the difference between the generated model distribution and the true distribution is minimized.

Consider a prior on input noise distribution $p_z(z)$. To learn the generator's distribution over data $x$, a mapping by the generator is performed from the noise distribution to data space as $G(z;\theta_g)$. We next define a discriminator $D(x;\theta_d)$ that outputs the probability whether $x$ comes from the generator or is real data. We try to train $G$ and $D$ simultaneously to minimize $log(1 - D(G(z))$ for $G$ and to minimize $log(D(X))$ for $D$. The standard formulation of GANs is given as follows:

\begin{equation}
\begin{split}
\min_G \max_D V (G, D) = E_{x \sim p_{data}(x)} [log D(x)] + \\ E_{z \sim p_z(z)}[log(1 - D(G(z)))]
\end{split}
\end{equation}
where $p_{data}$ is the data distribution and $p_z$ is the generator distribution to be learned through the adversarial min-max optimization.

\subsubsection{Conditional Adversarial Nets (CGANs)}
Mirza et al. \cite{cgan} developed Conditional-GANs based on the GANs described above. In a standard GAN, we cannot control the process of the data being generated. By conditioning the model on class labels, it is possible to direct the data generation process. Let $y$ be some extra information such as class label, then consider a conditional model where both the generator and discriminator are conditioned on $y$. 

The objective function of CGANs is as follows:

\begin{equation}
\begin{split}
\min_G \max_D V (G, D) = E_{x \sim p_{data}(x)} [log D(x|y)] + \\ E_{z \sim p_z(z)}[log(1 - D(G(z|y)))]
\end{split}
\end{equation}

The authors applied the Conditional-GANs framework on the MNIST dataset, and showed that it produced images that more closely resembled MNIST digits.

\subsubsection{SRCGAN}
Inspired by Conditional-GANs and SRGAN, we propose a GAN framework for image super-resolution with a class condition. We first downsample the HR images ($I^{HR}$) by a scale factor (i.e. x4) to create the LR input images ($I^{LR}$). Next, we feed the GAN with the LR images as well as the digit labels from the original images. This is the conditional element that we add to the GAN which makes our framework different. 

Our generator and discriminator are jointly conditioned as $D(x,y|\theta_d)$ and $G(x,y|\theta_g)$, respectively, where $x$ represents an HR image (either a true one or one produced by the generator) as the input, and $y$ is the class label. The objective of our adversarial model is:

\begin{equation}
\begin{split}
\min_G \max_D V (G, D) = E_{I^{HR} \sim  p_{data}(I^{HR})} [log (D(I^{HR},y))] + \\ E_{I^{LR}\sim p_G(I^{LR})}[log(1 - D(G(I^{LR},y),y))]
\end{split}
\end{equation}

\begin{figure}[h]
\centering
\includegraphics[scale=0.255]{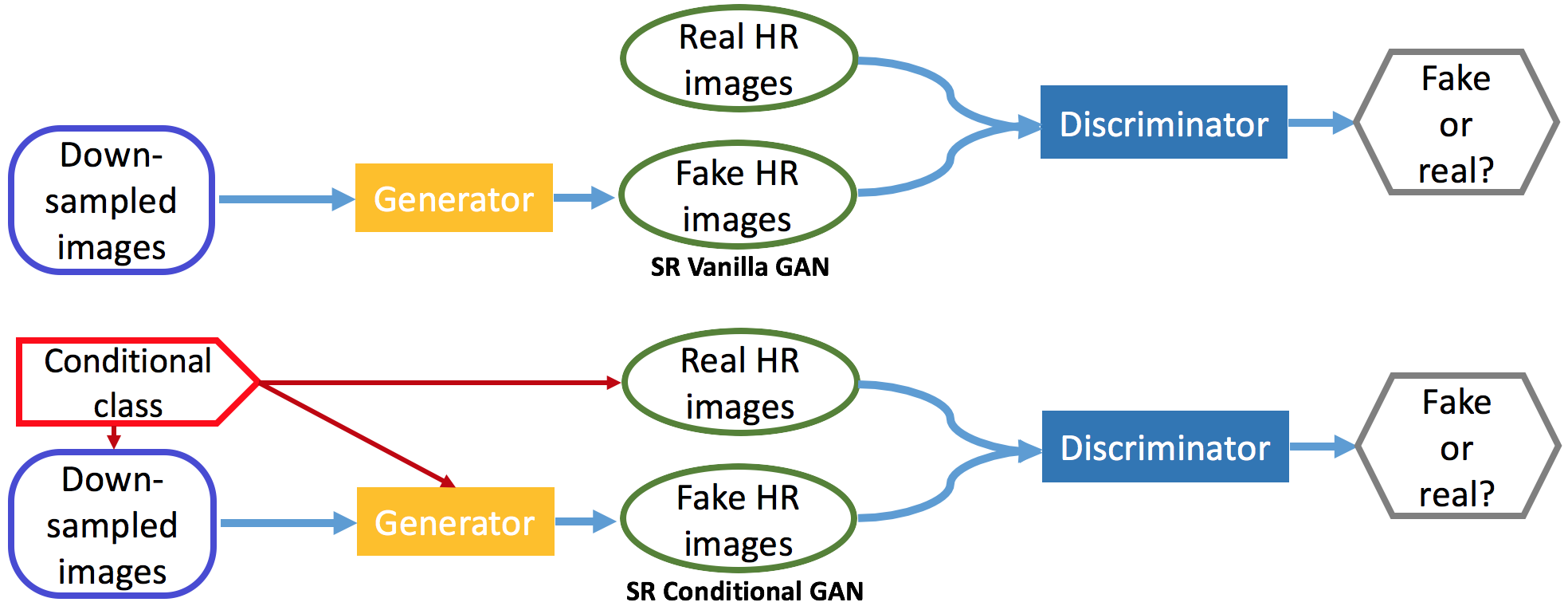}
\caption{(\textbf{top}) Vanilla GAN architecture. (\textbf{bottom}) Our proposed SRCGAN architecture. }
\end{figure}

\section{Experimental Results}

\subsection{VDSR-ResNeXt}
\subsubsection{Implementation details}
For training purposes, we used the same images as was used for training VDSR \cite{vdsr_paper} - 91 images from Yang et al. \cite{set91} and 200 images from Berkeley Segmentation Dataset \cite{set200_b100} - with data augmentation, and generated 41x41 patches from those images. The patches were then downsized by a factor of 2, 3 and 4, and upscaled to original dimensions using bicubic interpolation to be used as LR input for the network (370,000 images in total).

We also converted RGB-channel images to the YCbCr color space, and trained on the Y-channel, as was done by the authors of VDSR. This is because human vision is found to be more sensitive to details in intensity (measured by the Y-channel) than color(measured by Cb and Cr channels) \cite{ychannel}. 

We trained our model using a batch size of 128, stochastic gradient descent with momentum (momentum = 0.9, weight decay = 0.0001), with learning rate initialized at 0.1 and reduced by a factor of 10 every 10 epochs. In total, we trained all models for 25 epochs (216,675 iterations) due to the long training time required (~30min/epoch). Our model was written with the PyTorch library \cite{pytorch} and can support multiple Tesla P100 GPUs. 

\subsubsection{Quantitative results}
For validation purposes, we used images from Set5 \cite{set5}, Set14 \cite{set14} and B100 \cite{set200_b100}, which are well-established benchmark datasets for super-resolution tasks. 

We quantified the reconstruction accuracy of our model to ground truth using peak signal-to-noise ratio (PSNR) and structural similarity (SSIM), two widely used metrics for evaluating super-resolution results. As shown in Table 2, VDSR still recorded the highest PSNR and SSIM scores for most datasets and magnification factors, although our models come close, especially on the smaller magnification factor of x2, where our VDSR-ResNeXt-18-256 actually outperformed the original VDSR model.

\begin{table}[h]
\caption{PSNR/SSIM results for on Set5, Set14 and B100 for scale factors 2, 3 and 4}
\centering
 \begin{tabular}{|c|c|c|c|c|} 
 \hline
 Dataset & Scale & VDSR & Ours (18-128) & Ours (18-256) \\ [0.5ex] 
 \hline
 & x2 & 37.38/0.9587 & 37.29/0.9581 & 37.42/0.9588 \\
 Set5 & x3 & 33.57/0.9211 & 33.42/0.9195 & 33.55/0.9209 \\
 & x4 & 31.17/0.8804 & 30.98/0.8763 & 31.08/0.8785 \\
 \hline
 & x2 & 33.01/0.9128 & 32.95/0.9121 & 33.01/0.9132 \\
 Set14 & x3 & 29.80/0.8326 & 29.73/0.8308 & 29.79/0.8324 \\
 & x4 & 27.97/0.7663 & 27.82/0.7616 & 27.88/0.7637 \\
 \hline
 & x2 & 31.83/0.8956 & 31.79/0.8952 & 31.86/0.8963 \\
 B100 & x3 & 28.76/0.7977 & 28.69/0.7957 & 28.74/0.7975 \\
 & x4 & 27.15/0.7225 & 27.02/0.7173 & 27.06/0.7191 \\
 [1ex] 
 \hline
\end{tabular}
\end{table}
\setlength{\textfloatsep}{1pt plus 0.5pt minus 0.5pt}

\begin{figure}[h]
\centering
\includegraphics[scale=0.28]{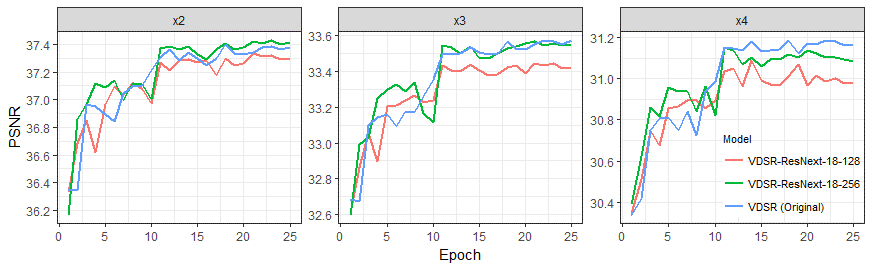}
\caption{PSNR on Set5 images during training on VDSR, VDSR-ResNeXt-18-128 and VDSR-ResNeXt-18-256. VDSR-ResNeXt-18-256 produces slightly better results than VDSR on the x2 scale.}
\end{figure}

\subsubsection{Qualitative results}
Fig. 4 shows a sample of results from our validation sets. The figure compares images generated from VDSR and our best proposed model against the LR input and HR ground truth. As shown, the predictions from our model have enhanced resolution compared to the LR input, although both VDSR and our model produce more blurry results in parts containing finer lines (e.g. eyelashes in the bottom image).

\begin{figure}[h]
\centering
\includegraphics[scale=0.225]{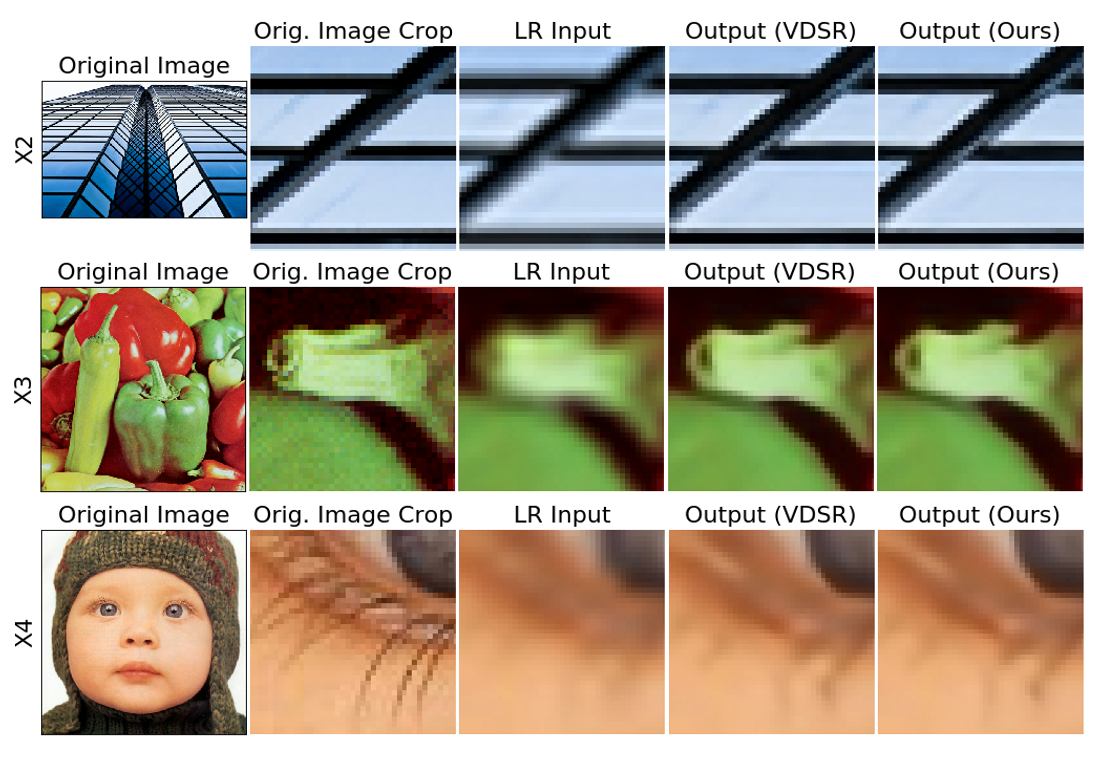}
\caption{Results for scale factors 2, 3 and 4 on three images from validation sets B100, Set14 and Set5, respectively. Our model (VDSR-ResNeXt-18-256) produces similar reconstruction results as VDSR.}
\end{figure}

\subsection{SRCGAN}

\subsubsection{Implementation details}
We used the MNIST \cite{MNIST} dataset, which contains 70,000 images (60,000 for training, 10,000 for testing) of handwritten digits from 0 to 9 with uniform digit distribution. The images are in greyscale with size $28\times 28$. We downscaled the images to $7\times7$ (factor of 4) with bicubic interpolation to use as input to the generator for our experiments (instead of random noise), along with the digit labels as the conditional variable. We assessed the performance of our proposed SRCGAN against a traditional GAN (i.e. without any conditional variables). 

SRCGAN was trained with a batch size of 128 and a learning rate of 0.001. The model was trained for 100 epochs (46,875 iterations). The leaky ReLU activation function and the Adam optimizer were used to build the model. The CNN classifier was trained with a batch size of 128 and a learning rate of 0.001. The model was trained for 70 epochs (32,813 iterations). The ReLU activation function and Adam optimizer were used in the model. Both models were written with the TensorFlow library \cite{tensor}.

\subsubsection{Quantitative results}
For evaluation purposes, we trained a CNN to classify digits of the MNIST dataset in order to assess whether predictions from our SRCGAN can be recognized as the correct digits. The classifier obtained an accuracy of 0.9861 on the MNIST test set. We fed generated images from our SRCGAN and vanilla GAN to the CNN classifier, and obtained classification accuracies of 0.8126 and 0.6811, respectively. This shows that the conditional information (digit in this case) helped us generate more accurate HR images.

\begin{table}[h]
\caption{MNIST digit classifier accuracy on images generated by SRCGAN and vanilla GAN.}
\centering
 \begin{tabular}{|c|c|c|} 
 \hline
  Models & SRCGAN & SR Vanilla GAN  \\ [1ex] 
 \hline
  Accuracy & 81.26\% & 68.11\%  \\
 [1ex] 
 \hline
\end{tabular}
\end{table}

\begin{figure}[h]
\centering
\includegraphics[scale=0.20]{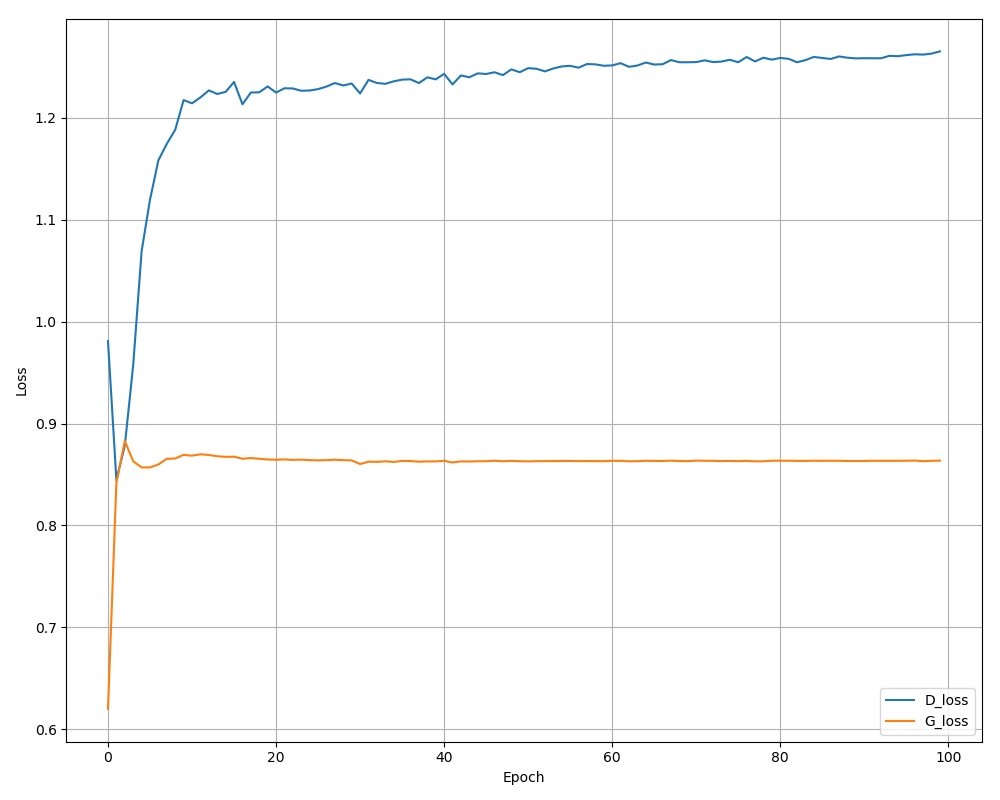}
\caption{Discriminator loss and the generator loss for SRCGAN during training.}
\end{figure}

\subsubsection{Qualitative results}
Fig. 6 compares the output from SRCGAN and vanilla GAN. Most of the images produced by the vanilla GAN are very blurry, making it difficult to tell what digits the images are supposed to represent. On the other hand, the images generated by SRCGAN are clearer and the digits can be easily recognized. The HR outputs of SRCGAN present obvious improvements over the downsized images. 

To sum up, comparisons based on both the CNN classifier and  visualization illustrate that the quality of HR image reconstruction can be improved by incorporating the conditional term.

\begin{figure}[h]
\centering
\includegraphics[scale=0.30]{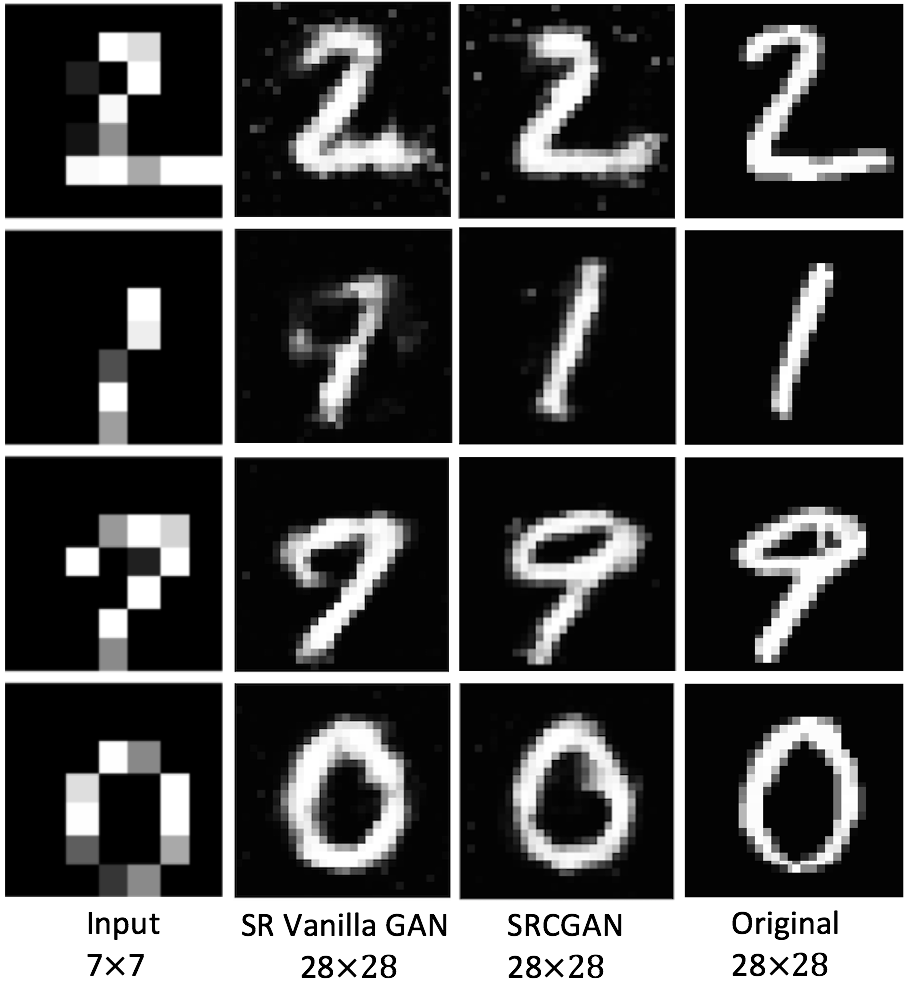}
\caption{MNIST classifier results on SRCGAN output and the vanilla GAN output, compared to the LR input and ground truth HR images.}
\end{figure}

\section{Conclusion and Future Work}
We proposed and implemented two approaches for the task of image super-resolution: VDSR-ResNeXt and SRCGAN.

\vspace{0.5em}

\textbf{VDSR-ResNeXt}: Even though we were unable to completely outperform the VDSR architecture, we believe there are several areas of future work that warrant further research. Firstly, we trained our model for 25 epochs due to computational time constraints while the original VDSR paper trained for 80 epochs, so there is room for further training, which could lead to small improvements in performance. Secondly, we kept the same settings (e.g. learning rate schedule) as the original VDSR model, which may not work well for the different architecture we proposed. Thirdly, we used PyTorch’s \cite{pytorch} built-in grouped convolution functionality, so there is potential for optimizing the code for parallel computations, which should reduce training time. We can also experiment with adding more residual connections, inspired by ResNet, to build even deeper structures. And lastly, having a greater variety of training data (e.g. ImageNet) should theoretically lead to better results for all models being compared. Our models lag behind VDSR as the scale factor increases, so we can also try supplying a greater proportion of training examples with the x4 scale factor.

\vspace{0.5em}

\textbf{SRCGAN}: After evaluating SRCGAN on the MNIST dataset, we successfully recovered clear HR images from the LR input. The model outputs are consistent with the ground truth MNIST images. Our model was able to produce more accurate reconstructed images compared to the vanilla GAN without conditional information. Moving forward, it would be interesting to try SRCGAN on more complex labelled datasets, such as CelebA \cite{celeba}, which contains about 200,000 celebrity faces along with 40 attribute annotations (e.g., gender, smiling, brown hair) that we can use as conditioning variables for our model.


%





\CLASSOPTIONcaptionsofftrue
\ifCLASSOPTIONcaptionsoff
  \newpage
\fi


\begin{thebibliography}{1}

\bibitem{tensor}
Abadi, Martín, and Barham, Paul, and Chen, Jianmin, and Chen, Zhifeng, and Davis, Andy, and Dean, Jeffrey, and Devin, Matthieu et al. "TensorFlow: A System for Large-Scale Machine Learning." In \emph{Symposium on Operating Systems Design and Implementation} vol. 16, pp. 265-283. 2016.

\bibitem{set5}
Bevilacqua, Marco, and Roumy, Aline, and Guillemot, Christine, and Morel, Marie-Line Alberi. “Low-Complexity Single-Image Super-Resolution Based on Nonnegative Neighbor Embedding.” In \emph{British Machine Vision Conference (BMVC)}, pp. 1-10, Sept. 2012.

\bibitem{srcnn_paper}
Dong, Chao,  and Loy, Chen Change, and He, Kaiming, and Tang, Xiaoou. “Image Super-Resolution Using Deep Convolutional Networks,” In \emph{IEEE Transactions on Pattern Analysis and Machine Intelligence}, vol. 38, no. 2, pp. 295–307, Feb. 2016.

\bibitem{GAN}
Goodfellow, Ian, and Pouget-Abadie, Jean, and Mirza, Mehdi, and Xu, Bing, and Warde-Farley, David, and Ozair, Sherjil, and Courville, Aaron, and Bengio, Yoshua. "Generative Adversarial Nets." In \emph{Advances in neural information processing systems,} pp. 2672-2680, 2014.

\bibitem{resnet_paper}
He, Kaiming, and Zhang, Xiangyu, and Ren, Shaoqing, and Sun, Jian. “Deep Residual Learning for Image Recognition.” In 2016 \emph{IEEE Conference on Computer Vision and Pattern Recognition (CVPR)}, pp. 770–778, June 2016.

\bibitem{Satellite_Imaging}
Jacobsen, Karsten. "High Resolution Satellite Imaging Systems - An Overview." In \emph{Photogrammetrie Fernerkundung Geoinformation} 2005, no. 6 (2005): 487.

\bibitem{vdsr_paper}
Kim, Jiwon, and Lee, Jung Kwon, and Lee, Kyoung Mu. "Accurate Image Super-Resolution Using Very Deep Convolutional Networks." In \emph{IEEE Conference on Computer Vision and Pattern Recognition (CVPR)}, June 2016.

\bibitem{MNIST}
LeCun, Yann. "The MNIST Database of Handwritten Digits." http://yann. lecun. com/exdb/mnist/ (1998).

\bibitem{srgan}
Ledig, Christian, and Theis, Lucas, and Huszár, Ferenc, and Caballero, Jose, and Cunningham, Andrew, and Acosta, Alejandro, and Aitken, Andrew, et al. "Photo-Realistic Single Image Super-Resolution Using a Generative Adversarial Network." arXiv preprint (2016).

\bibitem{celeba}
Liu, Ziwei, and Luo, Ping, and Wang, Xiaogang, and Tang, Xiaoou. "Deep Learning Face Attributes In the Wild." In \emph{Proceedings of International Conference on Computer Vision (ICCV)}, 2015.

\bibitem{set200_b100}
Martin, David, and Fowlkes, Charless, and Tal, Doron, and Malik, Jitendra. “A Database of Human Segmented Natural Images and Its Application to Evaluating Segmentation Algorithms and Measuring Ecological Statistics.” In \emph{Proc. 8th International Conference on Computer Vision}, vol. 2, pp. 416–423, July 2001.

\bibitem{cgan}
Mirza, Mehdi, and Osindero, Simon. "Conditional generative adversarial nets." arXiv preprint arXiv:1411.1784 (2014).

\bibitem{pytorch}
Paszke, Adam, and Gross, Sam, and Chintala, Soumith, and Chanan, Gregory, and Yang, Edward, and DeVito, Zachary, and Lin, Zeming, and Desmaison, Alban, and Antiga, Luca, and Lerer, Adam. "Automatic differentiation in PyTorch." In \emph{Conference on Neural Information Processing Systems (NIPS)}, Dec. 2017.

\bibitem{ychannel}
Podpora, Michal, and Korbas, Grzegorz Pawel, and Kawala-Janik, Aleksandra. "YUV vs RGB – Choosing a Color Space for Human-Machine Interaction." In 2014 \emph{Federated Conference on Computer Science and Information Systems}, pp. 29-34, Sept. 2014.

\bibitem{vgg}
Simonyan, Karen, and Zisserman, Andrew, “Very Deep Convolutional Betworks for Large-Scale Image Recognition.” In \emph{CoRR}, vol. abs/1409.1556, 2014.

\bibitem{compressed_image_improvement}
Tang, Jinshan, and Peli, Eli, and Acton, Scott. "Image enhancement using a contrast measure in the compressed domain." In \emph{IEEE Signal Processing Letters} 10, no. 10 (2003): 289-292.

\bibitem{facial_image}
Terzopoulos, Demetri, and Waters, Keith. "Analysis and Synthesis of Facial Image Sequences Using Physical and Anatomical Models." In \emph{IEEE Transactions on Pattern Analysis and Machine Intelligence} 15, no. 6 (1993): 569-579.

\bibitem{facial_image_enhancement}
Wang, Guoqiang, and Ou, Zongying. "Face Recognition Based on Image Enhancement and Gabor Features." In \emph{Intelligent Control and Automation}, 2006. WCICA 2006. The Sixth World Congress on, vol. 2, pp. 9761-9764. IEEE, 2006.

\bibitem{resnext_paper}
Xie, Saining, and Girshick, Ross, and Dollár, Piotr, and Tu, Zhuowen, and He, Kaiming. "Aggregated Residual Transformations for Deep Neural Networks." In \emph{IEEE Conference on Computer Vision and Pattern Recognition (CVPR)}, July 2017.

\bibitem{set91}
Yang, Jianchao, and Wright, John, and Huang, Thomas, and Ma, Yi. “Image Super-Resolution Via Sparse Representation,” In \emph{Transactions on Image Processing}, vol. 19, no. 11, pp. 2861–2873, Nov. 2010.

\bibitem{set14}
Zeyde, Roman, and Elad, Michael, and Protter, Matan. “On Single Image Scale-Up using Sparse-Representations.” In \emph{Curves \& Surfaces}, June 2010.

\bibitem{Med_image_processing}
Zikos, Marios, and Kaldoudi, Eleni, and Orphanoudakis, Stelios C.. "Medical Image Processing." In \emph{Student Health Technology and Informatics} 43 (1997): 465-469.

\end{thebibliography}
\end{document}